\documentclass[11pt,a4paper]{article}
\usepackage[hyperref]{emnlp2020}
\usepackage{times}
\usepackage{latexsym}
\usepackage{amsmath}
\usepackage{color}
\usepackage{bbm}
\usepackage[normalem]{ulem}

\usepackage{microtype}
\usepackage{booktabs}
\usepackage{multirow}
\usepackage{graphicx}
\aclfinalcopy 

\title{Leveraging Declarative Knowledge in Text and First-Order Logic \\for Fine-Grained Propaganda Detection}

\author{
	Ruize Wang$^{1}$\thanks{~~Work is done during internship at Microsoft Research Asia. }, 
	Duyu Tang$^2$, Nan Duan$^2$,Wanjun Zhong$^3$, Zhongyu Wei$^1$\thanks{{ }{ }Corresponding author.},\\
	\textbf{Xuanjing Huang$^1$, Daxin Jiang$^2$, Ming Zhou$^2$} \\
	$^1$Fudan University, Shanghai, China \\
	$^2$Microsoft Corporation, Beijing, China \\
	$^3$Sun Yat-Sen University, Guangzhou, China \\
    {\tt \{rzwang18,zywei,xjhuang\}@fudan.edu.cn}\\
    {\tt \{zhongwj25\}@mail2.sysu.edu.cn}\\
    {\tt \{dutang,nanduan,djiang,mingzhou\}@microsoft.com} 
}

\date{}

\begin{document}
\maketitle
\begin{abstract}
We study the detection of propagandistic text fragments in news articles.
Instead of merely learning from input-output datapoints in training data, we introduce an approach to inject declarative knowledge of fine-grained propaganda techniques. Specifically, we leverage the declarative knowledge expressed in both first-order logic and natural language.
The former refers to the logical consistency between coarse- and fine-grained predictions, which is used to regularize the training process with propositional Boolean expressions.
The latter refers to the literal definition of each propaganda technique, which is utilized to get class representations for regularizing the model parameters.
We conduct experiments on Propaganda Techniques Corpus, a large manually annotated dataset for fine-grained propaganda detection. 
Experiments show that our method achieves superior performance, demonstrating that leveraging declarative knowledge can help the model to make more accurate predictions.

\end{abstract}

\section{Introduction}

\begin{figure}[!t]
    \centering
    \includegraphics[width=1.0\linewidth]{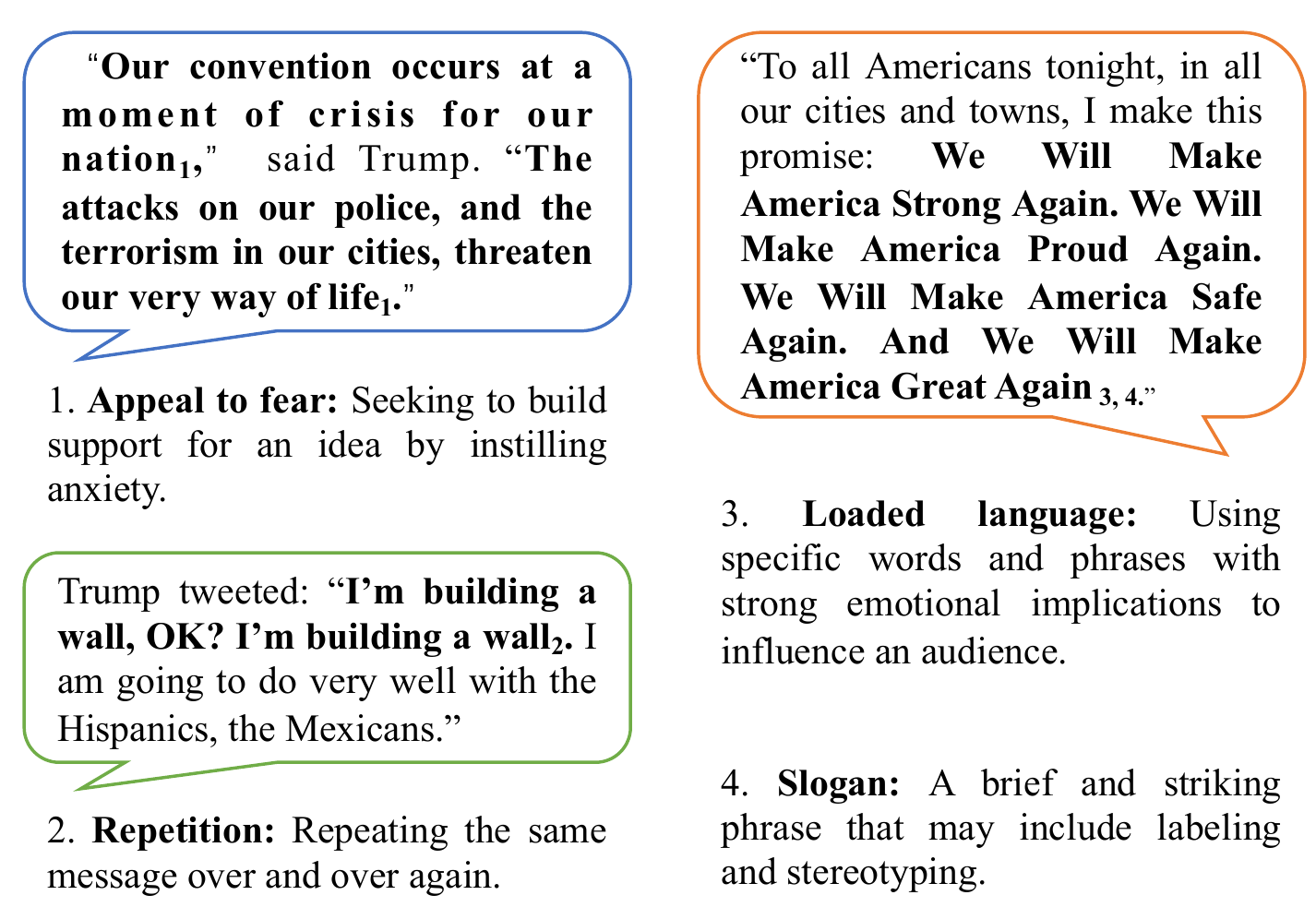}
    \caption{Examples of propagandistic texts, and definitions of corresponding propaganda techniques (\textbf{Bold} denotes propagandistic texts).  }
    \vspace*{-2.5ex}
    \label{fig:examples}
\end{figure}

Propaganda is the approach deliberately designed with specific purposes to influence the opinions of readers. Different from the fake news which is entirely made-up and has no verifiable facts, propaganda is possibly built upon an element of truth, and conveys information with strong emotion or somewhat biased. 
This characteristic makes propaganda more effective and unnoticed through the rise of social media platforms. 
Some examples of propagandistic texts and definitions of corresponding techniques are shown in Figure \ref{fig:examples}. 

We study the problem of fine-grained propaganda detection in this work, which is possible thanks to the recent release of Propaganda Techniques Corpus \citep{da2019fine}.
Different from earlier works \citep{rashkin2017truth,wang2017liar} that mainly study propaganda detection at a coarse-grained level, namely predicting whether a document is propagandistic or not, the fine-grained propaganda detection requires to identify the tokens of particular propaganda techniques in news articles.  
\citet{da2019fine} propose strong baselines in a multi-task learning manner, which are trained by binary detection of propaganda at sentence level and fine-grained propaganda detection over 18 techniques at token level. 
Such data-driven methods have the merits of convenient end-to-end learning and strong generalization, however, they cannot guarantee the consistency between sentence-level and token-level predictions. In addition, it is appealing to integrate human knowledge into data-driven approaches.

In this paper, we introduce an approach named \textbf{\textit{LatexPRO}} that leverages \textbf{\uline{l}}ogical \textbf{\uline{a}}nd \textbf{\uline{tex}}tual knowledge for \textbf{\uline{pro}}paganda detection.
Following \citet{da2019fine}, we develop a BERT-based multi-task learning approach as the base model, which makes predictions for 18 propaganda techniques at both sentence level and token level. 
Based on that, we inject two types of knowledge as additional objectives to regularize the learning process. 
Specifically, we exploit logic knowledge by transforming the consistency between sentence-level and token-level predictions with propositional Boolean expressions. Besides, we use the textual definition of propaganda techniques by first representing each of them as a contextual vector and then minimizing the distances to corresponding model parameters in semantic space. 

We conduct extensive experiments on Propaganda Techniques Corpus (PTC) \citep{da2019fine}, a large manually annotated dataset for fine-grained propaganda detection. 
Results show that our knowledge-augmented method significantly improves a strong multi-task learning approach.
In particular, our model greatly improves precision, demonstrating leveraging declarative knowledge expressed in both first-order logic and natural language can help the model to make more accurate predictions. More importantly, further analysis indicates that augmenting the learning process with declarative knowledge reduces the percentage of inconsistency in model predictions.

The contributions of this paper are summarized as follows:
\begin{itemize}
\item We introduce an approach to leverage declarative knowledge expressed in both first-order logic and natural language for fine-grained propaganda techniques.
\item We utilize both types of knowledge as regularizers in the learning process, which enables the model to make more consistent between sentence-level and token-level predictions.
\item Extensive experiments on the PTC dataset \citep{da2019fine} demonstrate that our method achieves superior performance with high $F_1$ and precision.
\end{itemize}

\begin{table}[]
\centering
\resizebox{\linewidth}{!}{%
\begin{tabular}{@{}llll@{}}
\toprule
\multirow{2}{*}{\textbf{Propaganda Technique}} & \multicolumn{3}{c}{\textbf{Instances}}        \\ \cmidrule(l){2-4} 
& \textbf{Train} & \textbf{Dev} & \textbf{Test} \\ \midrule
Loaded Language                                & 1,811           & 127          &  177             \\
Name Calling,Labeling                          & 931            & 68          &  86             \\
Repetition                                     & 456            & 35          &  80             \\
Doubt                                          & 423            & 23           &  44             \\
Exaggeration,Minimisation                      & 398            & 37           &  44             \\
Flag-Waving                                    & 206            & 13           &  21             \\
Appeal to fear-prejudice                       & 187            & 32           &  20             \\
Causal Oversimplification                      & 170            & 24           &  7             \\
Slogans                                        & 120            & 3           &  13             \\
Black-and-White Fallacy                        & 97             & 4           &  8             \\
Appeal to Authority                            & 91             & 2           &  23             \\
Thought-terminating Cliches                    & 70             & 4            &  5             \\
Whataboutism                                   & 55             & 1            &  1             \\
Reductio ad hitlerum                           & 44             & 5           &  5             \\
Red Herring                                    & 24             & 0            &    9           \\
Straw Men                                      & 11             & 0            &   2            \\
Obfus.,Int. Vagueness,Confusion                & 10             & 0            &  1             \\
Bandwagon                                      & 10             & 2            & 1              \\ \midrule
\textbf{Total}                                 & 5,114           & 380          & 547              \\ \bottomrule
\end{tabular}%
}
\caption{The statistics of all 18 propaganda techniques.}
\vspace*{-2.5ex}
\label{tab:statistics}
\end{table}

\section{Task}
\paragraph{Task Definition.}
Following the previous work \citep{da2019fine}, we conduct experiments on two different granularities tasks: sentence-level classification (SLC) and fragment-level classification (FLC).
Formally, in both tasks, the input is a plain-text document $d$. A document includes a set of propagandistic fragments $T$, in that each fragment is represented as a sequence of contiguous characters $t = [t_i,...,t_j]\subseteq d$. 
For the SLC task, the target is to predict whether a sentence is propagandistic which can be regarded as a binary classification problem.
For the FLC task, the target is to predict a set $S$ with propagandistic fragments $s = [s_m,...,s_n]\subseteq d$ and identify $s \in S$ to one of the propagandistic techniques.

\begin{figure*}[ht]
    \centering
    \includegraphics[width=0.85\linewidth]{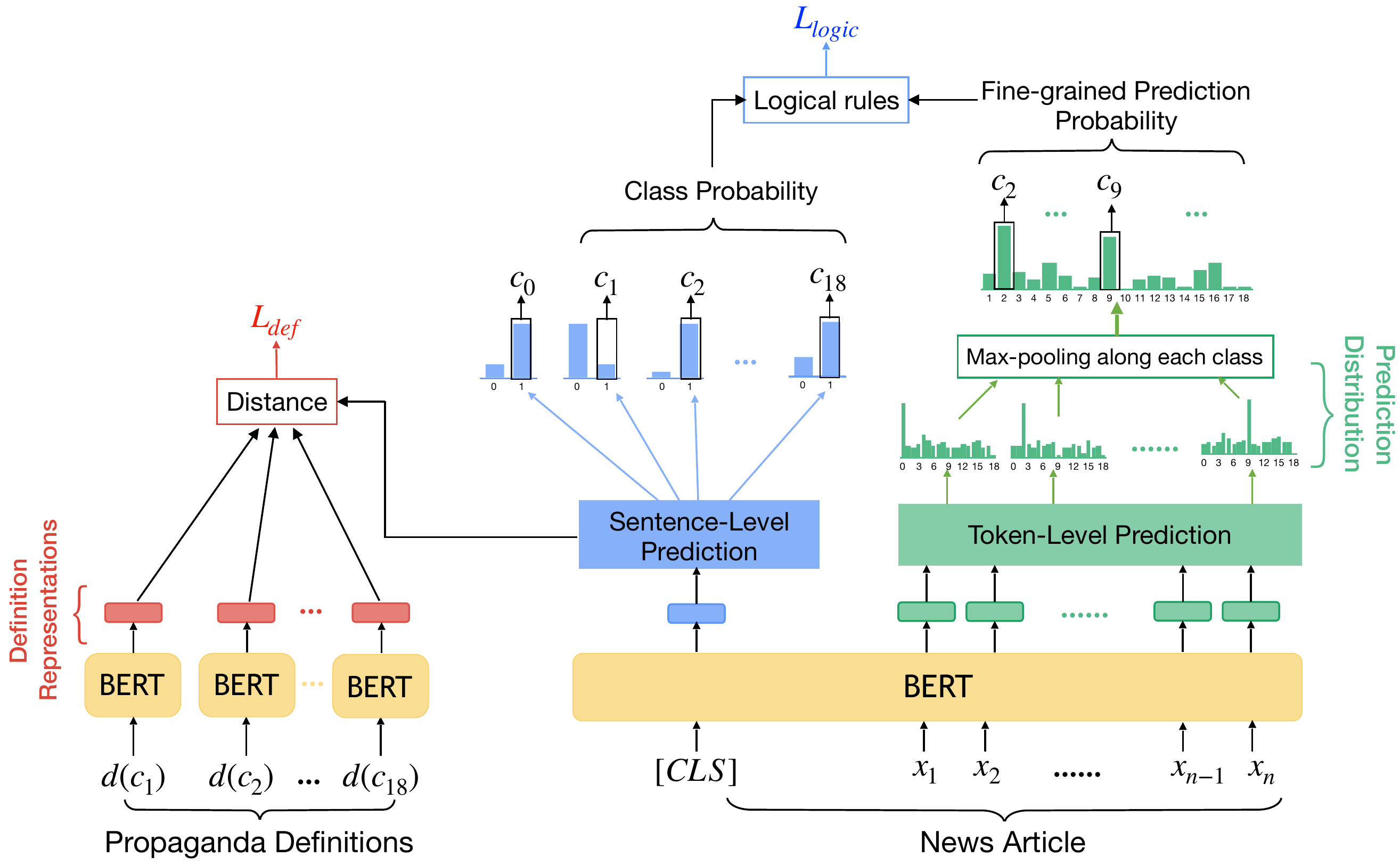}
    \caption{Overview of our proposed model. A BERT-based multi-task learning approach is adopted to make predictions for 18 propaganda techniques at both sentence level and token level. We introduce two types of knowledge as additional objectives: (1) logical knowledge about the consistency between sentence-level and token-level predictions, and (2) textual knowledge from literal definitions of propaganda techniques.
}
    \label{fig:overview}
\end{figure*}

\paragraph{Dataset.}
\label{Task_data}
This paper utilizes Propaganda Techniques Corpus (PTC) \citep{da2019fine} for experiments. PTC is a manually annotated dataset for fine-grained propaganda detection, containing 293/ 57/ 101 articles and 14,857/ 2,108/ 4,265 corresponding sentences for training, validation and testing, respectively. Each article is annotated with the start and end of the propaganda text span as well as the type of propaganda technique. As the annotations of the official testing set are not publicly available, we divided the official validation set into a validation set of 22 articles and a test set of 35 articles. The statistics of all 18 propaganda techniques and their frequencies (instances per technique) are shown as Table \ref{tab:statistics}.

\paragraph{Evaluation.}
For the SLC task, we evaluate the models with precision, recall and micro-averaged $F_1$ scores. As for the FLC task, we adopt the evaluation script provided by \citet{da2019fine} to calculate precision, recall, and micro-averaged $F_1$, in that giving partial credit to imperfect matches at the character level. The FLC task is evaluated on two kinds of measures: (1) \textit{Full task} is the overall task, which includes detecting the existence of propaganda techniques in text fragments and identifying the type of them, while (2) \textit{Spans} is a special case of the \textit{Full task}, which only considers the spans of propagandistic fragments except for their propaganda techniques.

\section{Method}
In this section, we present our approach \textit{LatexPRO} as shown in Figure \ref{fig:overview}, which injects declarative knowledge of fine-grained propaganda techniques into neural networks. We first present our base model (\S \ref{base_model}), which is a multi-task learning framework that slightly extends the model of \citet{da2019fine}. Afterward, we introduce two ways to regularize the learning process with logical knowledge about the consistency between sentence-level and token-level predictions (\S \ref{inject_logical_knowledge}) and textual knowledge from literal definitions of propaganda techniques (\S \ref{inject_textual_knowledge}). At last, we describe the training and inference procedures (\S \ref{training_inference}). 

\subsection{Base Model}
\label{base_model}
To better exploit the sentence-level information and further benefit token-level prediction, we develop a fine-grained multi-task method as our base model, which makes predictions for 18 propaganda techniques at both sentence level and token level.
Inspired by the success of pre-trained language models on various natural language processing downstream tasks, we adopt BERT \citep{devlin2018bert} as the backbone model here. For each input sentence, the sequence is modified as ``$[CLS]\textit{sentence tokens}[SEP]$''. 
Specifically, on top of BERT, we add 19 binary classifiers for fine-grained sentence-level predictions, and one 19-way classifier for token-level predictions, where all classifiers are implemented as linear layers. 
At sentence level, we perform multiple binary classifications and this can further support leveraging declarative knowledge. The last representation of the special token $[CLS]$ which is regarded as a summary of the semantic content of the input, is adopted to perform multiple binary classifications, including one binary classification to predict the existence of propaganda techniques, and 18 binary classifications to identify the types of them. We adopt sigmoid activation for each binary classifier.
At token level, the last representation of each token is fed into a linear layer to predict the propaganda technique over 19 categories (i.e., 18 categories of propaganda techniques plus one category for ``none of them''). We adopt Softmax activation for the 19-way classifier. Two different losses are applied for this multi-task learning process, including the sentence-level loss $L_{sen}$ and the token-level loss $L_{tok}$.
$L_{sen}$ is the binary cross-entropy loss of multiple binary classifications. $L_{tok}$ is the focal loss \citep{lin2017focal} of 19-way classification for each token, which could address the class imbalance problem.

\subsection{Inject Logical Knowledge}
\label{inject_logical_knowledge}
There are some implicit logical constraints between sentence-level and token-level predictions. 
However, neural networks are less interpretable and need to be trained with a large amount of data to make it possible to learn such implicit logic. Therefore, we consider tackling the problems by exploiting logic knowledge. In particular, we propose to employ propositional Boolean expressions to explicitly regularize the model with a logic-driven objective, which improves the logical consistency between two different grained predictions, and makes our method more interpretable.
For instance, in this work, if a propaganda class $c$ is predicted by the multiple binary classifiers (indicates the sentence contains this propaganda technique), then the token-level predictions belonging to the propaganda class $c$ should also exist. We thus consider the propositional rule $F=A \Rightarrow B$, formulated as:
\begin{equation}
\begin{aligned}
P(F) &=P(A \Rightarrow B) \\
&=\neg P(A) \vee P(B) \\
&=1-P(A)+P(A) P(B) \\
&=P(A)(P(B)-1)+1
\end{aligned}
\end{equation}
where A and B are two variables. Specifically, substituting $f_{c}(x)$ and $g_{c}(x)$ into above formula as $F=\forall c: f_{c}(x) \Rightarrow g_{c}(x)$, then the logic rule can be written as:
\begin{equation}
P(F) = P(f_{c}(x))(P(g_{c}(x))-1)+1
\end{equation}
where $x$ denotes the input, $f_{c}(x)$ is the binary classifier for the propaganda class $c$, and $g_{c}(x)$ is the probability of fine-grained predictions that contains $x$ being category of $c$. $g_{c}(x)$ can be obtained by max-pooling over all the probability of predictions for class $c$. Note that the probabilities of the unpredicted class are set to 0 to prevent any violation, i.e., ensuring that each class has a probability corresponding to it. Our objective here is maximizing $P(F)$, i.e., minimizing $L_{logic}=-log \left( P(F) \right)$, to improve the logical consistency between coarse- and fine-grained predictions.

\subsection{Inject Textual Knowledge}
\label{inject_textual_knowledge}
The literal definitions of propaganda techniques in this work, can be regarded as textual knowledge which contains useful semantic information. To exploit this kind of knowledge, we adopt an additional encoder to encode the literal definition of each propaganda technique. Specifically, for each definition, the input sequence is modified as ``$[CLS]\textit{definition}[SEP]$'' and fed into BERT. We adopt the last representation of the special token $[CLS]$ as each definition representation $D(c_i)$, where $c_i$ represents the $i$-th propaganda technique. We calculate the Euclidean distance $dist_2$ between each predicted propaganda category representation $W(c_i)$ and the definition representation $D(c_i)$. Our objective is minimizing the textual definition loss $L_{def}$, which regularizes the model to refine the propaganda representations.
\begin{equation}
L_{def} = \sum_{i=1}^{18}dist_2\left(W(c_i),D(c_i)\right)
\end{equation}

\subsection{Training and Inference}
\label{training_inference}
\paragraph{Training.}
To train the whole model jointly, we introduce a weighted sum of losses $L_j$ which consists of the token-level loss $L_{tok}$, fine-grained sentence-level loss $L_{sen}$, textual definition loss $L_{def}$ and logical loss $L_{logic}$:
\begin{equation}
\label{loss}
L_j=\alpha* L_{tok} + \beta*\left(L_{sen}+L_{def} * \lambda\right) + \gamma* L_{logic}
\end{equation}
where hyper-parameters $\alpha$, $\beta$, $\lambda$ and $\gamma$ are employed to control the tradeoff among losses. During the training stage, our goal is minimizing $L_j$.

\begin{table*}[!t]
\centering
\begin{tabular}{@{}lccc|cccc@{}}
\toprule
\multirow{2}{*}{Model}     & \multicolumn{3}{c|}{Spans} & \multicolumn{4}{c}{Full Task} \\ \cmidrule(l){2-8} 
 & \multicolumn{1}{c}{P} & \multicolumn{1}{c}{R} & \multicolumn{1}{c|}{$\text{F}_1$} & \multicolumn{1}{c}{P} & \multicolumn{1}{c}{R} & \multicolumn{1}{c|}{$\text{F}_1$}
 & \multicolumn{1}{c}{$M_C$} \\ 
 \midrule
BERT \citep{da2019fine}                  & 50.39   & 46.09  & 48.15  & 27.92    & 27.27    & \multicolumn{1}{c|}{27.60} & -   \\
MGN \citep{da2019fine}                   & 51.16   & \bf{47.27}  & 49.14  & 30.10    & \bf{29.37}    & \multicolumn{1}{c|}{29.73} & -   \\
\midrule
LatexPRO            & 58.95   & 42.37  & 49.30  &40.98    & 26.99    & \multicolumn{1}{c|}{32.54}  & 16.05  \\
LatexPRO (L)           & \bf{61.61}   & 43.41  & 50.93  & 42.44    & 28.25    & \multicolumn{1}{c|}{33.92} & 21.86  \\
LatexPRO (T)           & 61.20   & 42.67  & 50.28 &  41.91    & 28.06    & \multicolumn{1}{c|}{33.61} & 19.29  \\
LatexPRO (L+T) & 61.22   & 45.18  & \bf{51.99}  & \bf{42.64}    & 29.17    & \multicolumn{1}{c|}{\bf{34.65}}   & \bf{23.62}
\\ \bottomrule
\end{tabular}%

\caption{Overall performance on fragment-level experiments (FLC task) in terms of Precision (P), recall (R) and $\text{F}_1$ scores on our test set. $M_C$ denotes the metric of consistency between sentence-level predictions and token-level predictions. \textit{Full task} is the overall task of detecting both propagandistic fragments and identifying the technique, while \textit{Spans} is a special case of the \textit{Full task}, which only considers the spans of fragments except for their propaganda techniques. Note that (L+T), (L), and (T) denote injecting of both logical and textual  knowledge, only logical knowledge, and only textual knowledge, respectively. }
\label{tab:flc}
\end{table*}

\paragraph{Inference.}
For the SLC task, our method output the ``\textit{propaganda}'' only if the probability of propagandistic binary classification for the positive class is above 0.7. This threshold is chosen according to the ratio of propaganda to non-propaganda samples in the training set.
For the FLC task, to better exploit the coarse-grained (sentence-level) information to guide the fine-grained (token-level) prediction, we design a way that can explicitly make constraints on 19-way predictions when doing inference. Prediction probabilities of 18 fine-grained binary classifications above 0.9 are set to 1, and vice versa to 0. Then the Softmax probability of 19-way predictions (except for the ``\textit{none of them}'' class) of each token is multiplied by the corresponding 18 probabilities of propaganda techniques. 
This means that our model is conservative, which makes predictions for the fragments of propaganda techniques only if with high confidence.

\section{Experiments}
\subsection{Experimental Settings}
In this paper, we conduct experiments on Propaganda Techniques Corpus (PTC)\footnote{Note that the annotations of the official PTC test set are not publicly available, thus we split the original dev set into dev and test set as Section \ref{Task_data}. We use the released code \cite{da2019fine} to run the baseline.} \citep{da2019fine} which is a large manually annotated dataset for fine-grained propaganda detection, as detailed in Section 2.  
$F_1$ score is adopted as the final metric to represent the overall performance of models.
We select the best model on the dev set. 

We adopt BERT-base-cased \citep{devlin2018bert} as the pre-trained model. We implement our model using Huggingface \citep{Wolf2019HuggingFacesTS}. We use AdamW as the optimizer. 
In our best model on the dev set, the hyper-parameters in loss optimization are set as $\alpha = 0.8$, $\beta=0.2$, $\lambda=0.001$ and $\gamma=0.001$. We set the max sequence length to 256, the batch size to 16, the learning rate to 3e-5 and warmup steps to 500. We train our model for 20 epochs and adopt an early stopping strategy on the average validation $F_1$ score of \textit{Spans} and \textit{Full Task} with patience of 5 
epochs. For all experiments, we set the random seed to 42 for reproducibility. 


\subsection{Models for Comparison}
We compare our proposed methods with several baselines. Moreover, three variants of our method are provided to reveal the impact of each component.
The notations of LatexPRO (L+T), LatexPRO (L), and LatexPRO (T) denote our model which injects both logical and textual knowledge, only logical knowledge and only textual knowledge, respectively. 
Each of these models are described as follows.

\textbf{BERT \citep{da2019fine}} adds a linear layer on the top of BERT, and is fine-tuned on the SLC and FLC tasks, respectively.

\textbf{MGN \citep{da2019fine}} is a multi-task learning model, which regards the SLC task as the main task and drive the FLC task on the basis of the SLC task.

\textbf{LatexPRO} is our base model without leveraging any declarative knowledge.

\textbf{LatexPRO (L)} injects logical knowledge into \textit{LatexPRO} by employing propositional Boolean expressions to explicitly regularize the model.

\textbf{LatexPRO (T)} arguments \textit{LatexPRO} with textual knowledge in the literal definitions of propaganda techniques.

\textbf{LatexPRO (L+T)} is our full model in this paper.

\begin{table*}[ht]
\small
\centering
\resizebox{\textwidth}{!}{%
\begin{tabular}{@{}llll|lll|lll@{}}
\toprule
\multirow{2}{*}{\textbf{Propaganda Technique}} &
  \multicolumn{3}{c|}{\textbf{MGN}} &
  \multicolumn{3}{c|}{\textbf{LatexPRO}} &
  \multicolumn{3}{c}{\textbf{LatexPRO (L+T)}} \\ \cmidrule(l){2-10} 
 &
  \multicolumn{1}{c}{\textbf{P}} &
  \multicolumn{1}{c}{\textbf{R}} &
  \multicolumn{1}{c|}{\textbf{$\text{F}_1$}} &
  \multicolumn{1}{c}{\textbf{P}} &
  \multicolumn{1}{c}{\textbf{R}} &
  \multicolumn{1}{c|}{\textbf{$\text{F}_1$}} &
  \multicolumn{1}{c}{\textbf{P}} &
  \multicolumn{1}{c}{\textbf{R}} &
  \multicolumn{1}{c}{\textbf{$\text{F}_1$}} \\ \midrule
Appeal to Authority             &0  &0  &0  &0  &0  &0  & 0 &0  & 0  \\
Appeal to fear-prejudice        &8.41  &18.26  &11.52  &15.69  &14.90  &15.28  & 13.53 &14.90  & 14.18  \\
Bandwagon                       &0  &0  &0  & 0 &0  &0  & 0 &0  & 0  \\
Black-and-White Fallacy         &31.97  &43.12  &36.72  & 66.67 & 7.23 &13.05  & 81.63 &15.04  &25.41  \\
Causal Oversimplification       &12.43  &12.09  &12.66  & 12.43 & 30.00 &17.59  & 16.53 &28.57  & 20.94 \\
Doubt                           &27.12  &12.38  &17.00  & 18.06 & 9.09 &12.09  & 40.82 & 9.26 & 15.10 \\
Exaggeration,Minimisation       &33.95  &11.94  &17.67  & 42.85 &5.86  &10.31  & 31.57 &8.56  &13.47  \\
Flag-Waving                     &45.61  &37.71  &41.29  & 44.18 &36.13  &39.75  & 35.16 &41.30  &37.98  \\
Loaded Language                 &37.20  &46.45  &41.31  & 51.69 & 39.19 & 44.58 & 50.28 &44.39  &47.15  \\
Name Calling,Labeling           &36.15  &25.86  &30.15  &38.87  & 29.14 &33.31  & 43.09 &31.12  & 36.14  \\
Obfus.,Int. Vagueness,Confusion & 0  &0  &0  & 100.00 &98.61  &99.30  & 50.00 &98.61  &66.35  \\
Red Herring                     &0  &0  &0  &0  &0  & 0 & 0 & 0  & 0  \\
Reductio ad hitlerum            &45.40  &49.02  &47.14  &99.85  &59.88  &74.87  & 100.00 &45.74  &62.77  \\
Repetition                      &35.05  &24.09  &26.93  & 46.06 &28.75  &35.40  & 48.24 &26.86  &34.51  \\
Slogans                         &30.10  &31.25  &30.66  & 44.30 &38.46  &41.17  & 41.53 & 43.43  & 42.46  \\
Straw Men                       &0  &0  &0  &0  &0  &0  & 0 &0  &0  \\
Thought-terminating Cliches     &21.05  &23.85  &22.36  & 90.83 &14.80  &25.45  & 89.49 &19.60  &32.16  \\
Whataboutism                    &0  &0  & 0 & 9.09 &66.50  &15.99  & 18.75 &14.50  &16.35  \\ \bottomrule
\end{tabular}%
}
\caption{Detailed performance on the full task of fragment-level experiments (FLC task) on our test set. Precision (P), recall (R) and $\text{F}_1$ scores per technique are provided.}
\label{tab:flc2}
\vspace{-2mm}
\end{table*}

\subsection{Experiment Results and Analysis}
\paragraph{Fragment-Level Propaganda Detection.}
The results for the FLC task are shown in Table \ref{tab:flc}.
Our base model \textit{LatexPRO} achieves better results than other baseline models, which verifies the effectiveness of our fine-grained multi-task learning structure. 
It is worth noting that, our full model \textit{LatexPRO (L+T)} achieves superior boost than \textit{MGN} by 10.06\% precision and 2.85\% $F_1$ on the \textit{Spans} task, 12.54\% precision and 4.92\% $F_1$ on the \textit{Full} task, which is considered as significant progress. This demonstrates that leveraging declarative knowledge in text and first-order logic helps to predict the propaganda types more accurately. Moreover, our ablated models \textit{LatexPRO (T)} and \textit{LatexPRO (L)} both gain improvements over \textit{LatexPRO}, while \textit{LatexPRO (L)} gains more improvements than \textit{LatexPRO (T)}. This indicates that injecting each kind of knowledge is useful, and the effect of different kinds of knowledge can be superimposed and uncoupled.
It should be noted that, compared with baseline models, our models achieve a superior performance thanks to high precision, but the recall slightly loses. This is mainly because our models tend to make predictions for the high confident propaganda types.

To further understand the performance of models for the FLC task, we make a more detailed analysis of each propaganda technique. Table \ref{tab:flc2} shows detailed performance on the \textit{Full} task. Our models achieve precision and $F_1$ improvements of almost all the classes over baseline model, and can also predict some low-frequency propaganda techniques, e.g., \texttt{Whataboutism} and \texttt{Obfus.,Int}.
This further demonstrates that our method can stress class imbalance problem, and make more accurate predictions.

\begin{table}[!t]
\centering
\resizebox{\linewidth}{!}{%
\begin{tabular}{@{}llll@{}}
\toprule
Model                       & P & R & $\text{F}_1$ \\ \midrule
Random		&30.48   &51.04   &38.16    \\
All-Propaganda 	&30.54   &100.00   &46.80    \\
\midrule
BERT \citep{da2019fine}		&58.26   &57.81   &58.03    \\
MGN \citep{da2019fine}		&57.41   &62.50   &59.85    \\
\midrule
LatexPRO                    &56.18   &69.79   &62.25    \\
LatexPRO (L)          &56.53   &\bf{73.17}   &63.79    \\
LatexPRO (T)            &58.33   &67.50   &62.58    \\
LatexPRO (L+T) &\bf{59.04}   &71.66   &\bf{64.74}    \\ \bottomrule
\end{tabular}%
}
\caption{Results on sentence-level experiments (SLC task) in terms of Precision (P), recall (R) and $\text{F}_1$ scores on our test set. \textit{Random} is a baseline which predicts randomly, and \textit{All-Propaganda} is a baseline always predicts the propaganda class.}
\label{tab:slc}
\vspace{-3mm}
\end{table}

\paragraph{Sentence-Level Propaganda Detection.}
Table \ref{tab:slc} shows the performances of different models for SLC. The results indicate that our model achieves superior performances over other baseline models. Compared with \textit{MGN}, our \textit{LatexPRO (L+T)} increases the precision by 1.63\%, recall by 9.16\% and $F_1$ score by 4.89\%. This demonstrates the effectiveness of our model, and shows that our model can find more positive samples which will further benefit the token-level predictions for FLC.

\begin{figure}[!t]
    \centering
    \includegraphics[width=1.0\linewidth]{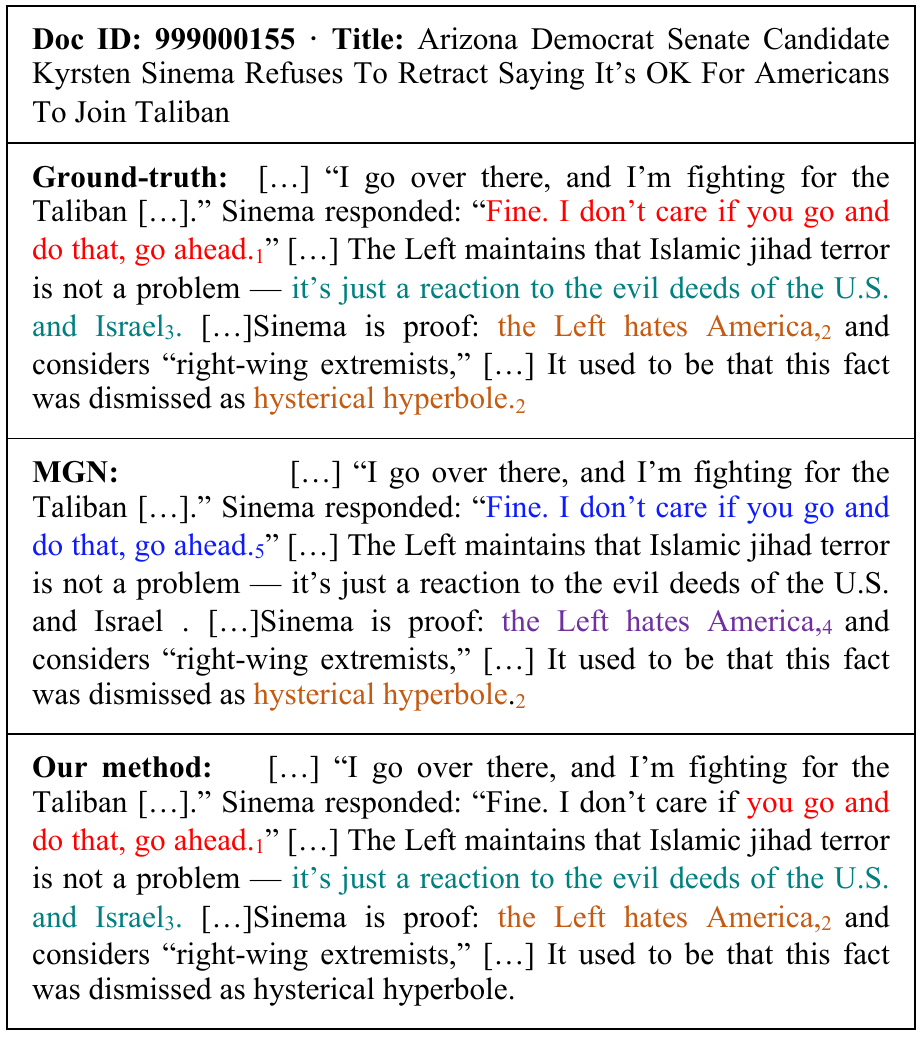}
    \caption{
    Qualitative comparison of 2 different models on a news article. The baseline \textit{MGN} predicts spans of fragments with wrong propaganda techniques, while our method can make more accurate predictions. 
    Here are 5 propaganda techniques:
    \textcolor{red}{\texttt{1.Thought-terminating Cliches}}, \textcolor{brown}{\texttt{2.Loaded Language}}, \textcolor{teal}{\texttt{3.Causal Oversimplification}}, \textcolor{violet}{\texttt{4.Flag waving}} and \textcolor{blue}{\texttt{5.Repetition}}. (Best viewed in color)
    }
    \label{fig:case_study}
    \vspace{-3mm}
\end{figure}

\subsection{Effectiveness of Improving Consistency}
We further define the following metric $M_C$ to measure the consistency between sentence-level predictions $Y_c$ which is a set of predicted propaganda technique classes, and token-level predictions $Y_t$ which is a set of predicted propaganda techniques for input tokens:
\begin{equation}
M_C(Y_c, Y_t)=\frac{1}{|Y_t|} \sum_{y_t \in Y_t} \mathbbm{1}_{Y_c}(y_t)
\end{equation}
where $|Y_t|$ denotes a normalizing factor, $\mathbbm{1}_{A}(x)$ represents the indicator function: 
\begin{equation}
\mathbbm{1}_{A}(x)=\left\{\begin{array}{ll}
1 & \text { if } x \in A \\
0 & \text { if } x \notin A
\end{array}\right.
\end{equation}
Table \ref{tab:flc} presents the consistency scores $M_C$. The higher the score indicates the better consistency. 
Results illustrate that our methods with declarative knowledge can substantially outperform the base model \textit{LatexPRO}. 
Compared to the base model, our declarative-knowledge-augmented methods enrich the source information by introducing textual knowledge from propaganda definitions, and logical knowledge from implicit logical rules between predictions, which enables the model to make more consistent predictions.

\begin{figure}[!t]
    \centering
    \includegraphics[width=1.0\linewidth]{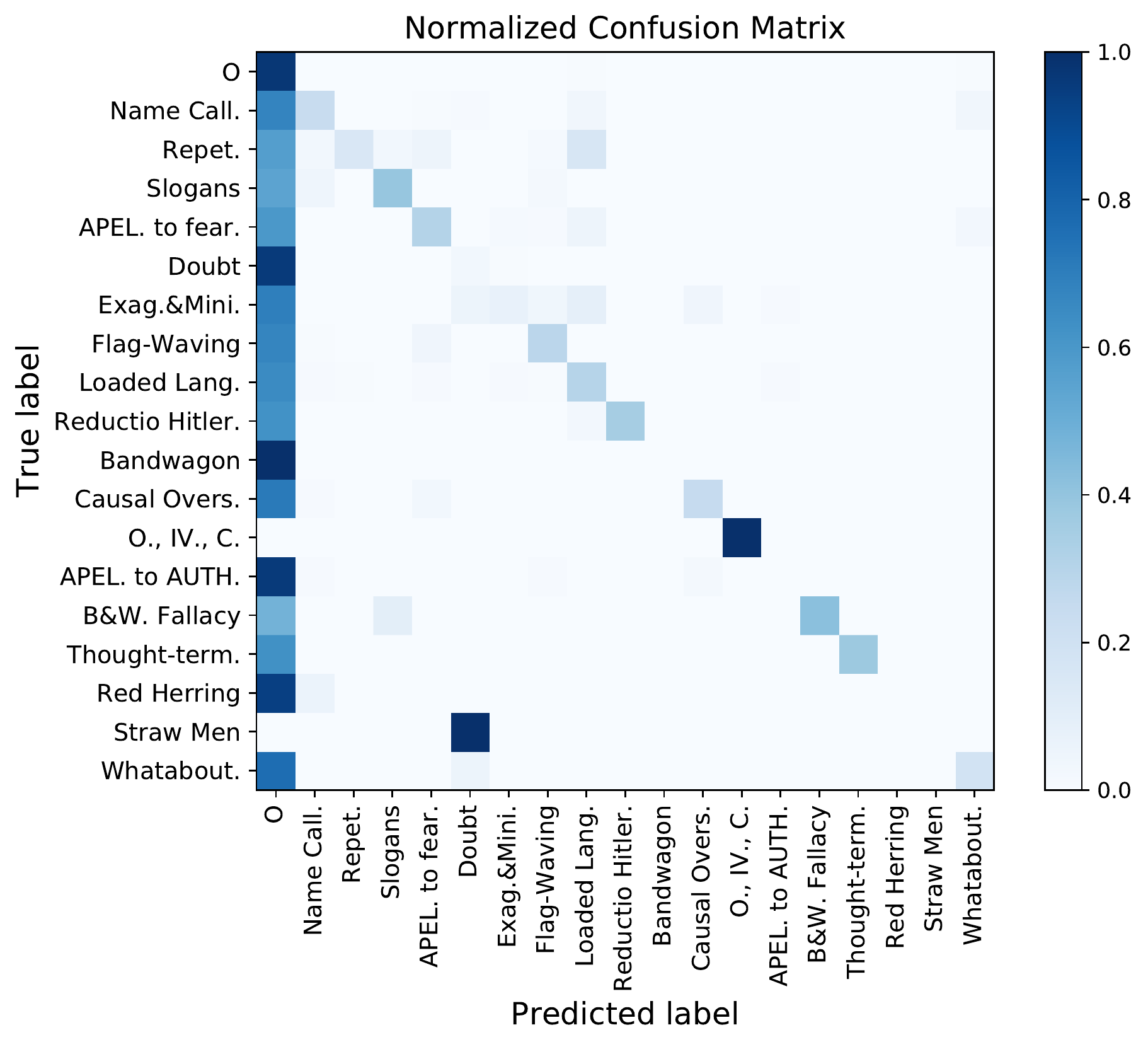}
    \caption{Visualization of confusion matrix result of our \textit{LatexPRO (L+T)}, where \texttt{O} represents the \texttt{none of them} class.}
    \label{fig:confusion_mat1}
    \vspace{-4mm}
\end{figure}

\subsection{Case Study}
Figure \ref{fig:case_study} gives a qualitative comparison example between \textit{MGN} and our \textit{LatexPRO (L+T)}. Different colors represent different propaganda techniques. The results show that although \textit{MGN} could predict the spans of fragments correctly, it fails to identify their techniques to some extent. However, our method shows promising results on both spans and specific propaganda techniques, which further confirms that our method can make more accurate predictions.

\subsection{Error Analysis}
Although our model has achieved the best performance, it still some types of propaganda techniques are not identified, e.g., \texttt{Appeal to Authority} and \texttt{Red Herring} as shown in Table \ref{tab:flc2}. To explore why our model \textit{LatexPRO (L+T)} cannot predict for those propaganda techniques, we compute a confusion matrix for the \textit{Full Task} of FLC task, and visualize the confusion matrix using a heatmap as shown in Figure \ref{fig:confusion_mat1}. We find that most of the off-diagonal elements are in class \texttt{O} which represents \texttt{none of them}. This demonstrates most of the cases are wrongly classified into \texttt{O}. We think this is due to the imbalance of the \texttt{propaganda} and \texttt{non-propaganda} categories in the dataset.
Similarly, \texttt{Straw Men}, \texttt{Red Herring} and \texttt{Whataboutism} are the relatively low frequency of classes.
How to deal with the class imbalance still needs further exploration.

\section{Related work}
Our work relates to fake news detection and the injection of first-order logic into neural networks. We will describe related studies in these two directions.

Fake news detection draws growing attention as the spread of misinformation on social media becomes easier and leads to stronger influence. Various types of fake news detection problems are introduced.
For example, there are 4-way classification of news documents \cite{rashkin2017truth}, and 6-way classification of short statements \cite{wang2017liar}.
There are also sentence-level fact checking problems with various genres of evidence, including natural language sentences from Wikipedia \cite{thorne2018fever}, semi-structured tables \cite{chen2019tabfact}, and images \cite{zlatkova2019fact,nakamura2019r}. 
Our work studies propaganda detection, a fine-grained problem that requires token-level prediction over 18 fine-grained propaganda techniques. The release of a large manually annotated dataset \cite{da2019fine} makes the development of large neural models possible, and also triggers our work, which improves a standard multi-task learning approach by augmenting declarative knowledge expressed in both first-order logic and natural language.

Neural networks have the merits of convenient end-to-end training and good generalization, however, they typically need a lot of training data and are not interpretable. On the other hand, logic-based expert systems are interpretable and require less or no training data. It is appealing to leverage the advantages from both worlds. In NLP community, the injection of logic to neural network can be generally divided into two groups. 
Methods in the \textbf{first} group regularize neural network with logic-driven loss functions \cite{xu2017semantic,fischer2018dl2,li2019logic}.
For example, \newcite{rocktaschel2015injecting} target on the problem of knowledge base completion. After extracting and annotating propositional logical rules about relations in knowledge graph, they ground these rules to facts from knowledge graph and add a differentiable training loss function. 
\newcite{kruszewski2015deriving} map text to Boolean representations, and derive loss functions based on implication at Boolean level for entailment detection. 
\newcite{demeester2016lifted} propose lifted regularization for knowledge base completion to improve the logical loss functions to be independent of the number of grounded instances and to further extend to unseen constants,  
The basic idea is that hypernyms have ordering relations and such relations correspond to component-wise comparison in semantic vector space.  
\newcite{hu2016harnessing} introduce a teacher-student model, where the teacher model is a rule-regularized neural network, whose predictions are used to teach the student model. 
\newcite{wang2018deep} generalize virtual evidence \cite{pearl2014probabilistic} to arbitrary potential functions over inputs and outputs, and use deep probabilistic logic to integrate indirection supervision into neural networks.  
More recently, \newcite{Asai2020LogicGuidedDA} regularize question answering systems with symmetric consistency  and symmetric consistency. The former creates a symmetric question by replacing words with their antonyms in comparison question, while the latter is for causal reasoning questions through creating new examples when positive causal relationship between two cause-effect questions holds. 

The \textbf{second} group is to incorporate logic-specific modules into the inference process \cite{yang2017differentiable,dong2019neural}. For example, \newcite{rocktaschel2017end} target at the problem of knowledge base completion, and use neural unification modules to recursively construct model similar to the backward chaining algorithm of Prolog. \newcite{evans2018learning} develop a differentiable model of forward chaining inference, where weights represent a probability distribution over clauses. \citet{li2019augmenting} inject logic-driven neurons to existing neural networks by measuring the degree of the head being true measured by probabilistic soft logic \citep{kimmig2012short}. Our approach belongs to the first direction, and to the best of knowledge our work is the first one that augments neural network with logical knowledge for propaganda detection. 

\section{Conclusion}
In this paper, we propose a fine-grained multi-task learning approach, which leverages declarative knowledge to detect propaganda techniques in news articles. Specifically, the declarative knowledge is expressed in both first-order logic and natural language, which are used as regularizers to obtain better propaganda representations and improve logical consistency between coarse- and fine-grained predictions, respectively.
Extensive experiments on the PTC dataset demonstrate that our knowledge-augmented method achieves superior performance with more consistent between sentence-level and token-level predictions.

\section*{Acknowledgments}
This work is partically supported by National Natural Science Foundation of China (No. 71991471), Science and Technology Commission of Shanghai Municipality Grant (No.20dz1200600, No.18DZ1201000, 17JC1420200).

\bibliography{emnlp2020}
\bibliographystyle{acl_natbib}

\end{document}